\documentclass[10pt,twocolumn,letterpaper]{article}

\usepackage{wacv}

\usepackage{multirow}
\usepackage{array}
\usepackage{tikz}
\usetikzlibrary{positioning,arrows.meta,shapes.geometric,calc}

\newtheorem{lemma}{Lemma}
\newtheorem{corollary}{Corollary}
\newcommand{\yhat}{\hat{y}}

\definecolor{wacvblue}{rgb}{0.21,0.49,0.74}
\usepackage[pagebackref,breaklinks,colorlinks,allcolors=wacvblue]{hyperref}

\def\wacvPaperID{*****} 
\def\confName{WACV}
\def\confYear{2027}

\title{Answer or Defer? Testing VLMs’ Stated Confidence on Brain MRI}

\author{
Amir Sabbaghziarani$^{1,2}$ \quad
Parsa Ghazvinian$^{2}$ \quad
Sepehr Salem$^{2}$ \quad
Sergey Plis$^{1,2}$ \\
\\
$^{1}$ Joint Georgia Institute of Technology / Georgia State University / Emory University\\
Center for Translational Research in Data Science and Neuroimaging\\
$^{2}$ Georgia State University
}

\begin{document}
\maketitle

\begin{abstract}
Vision--language models (VLMs) could support medical-image interpretation, but their use requires deciding which interpretations to accept and which images to refer to a clinician. We test whether thresholding stated confidence can retain useful coverage while limiting errors among accepted answers, and how this depends on the population. Eight open-weight VLMs, including two reasoning models, assess 3,888 brain MRI slices from post-treatment glioma patients (BraTS 2024) and healthy volunteers (IXI) for tumor or gross abnormality. BraTS labels are mask-derived; IXI slices are treated as negatives. On BraTS, no threshold selects a nonempty set with at most 5\% empirical error for any model, even when chosen retrospectively. This also holds for four models evaluated on a large-lesion, distant-negative subset. At a 10\% error budget, Gemma-4-12B retains 24.4\% of slices and every other model at most 2.5\%. To investigate this limited selection ability, we separate confidence differences between predicted answers from differences within each answer. For four models with available outcome-level summaries, pooled mean confidence differs more between answers than between correct and incorrect responses within each answer. A covariance decomposition explains how this pattern can produce opposite confidence--correctness associations across populations. At a 20\% error budget, accepting every ``tumor'' answer covers more slices than any confidence threshold for two models. These findings caution against using unvalidated stated confidence as a standalone deferral signal and support population-specific validation before deployment. Code is available at https://github.com/amir-sbg/VLM-Answer-or-Defer.
\end{abstract}
\begin{figure*}[t]
\centering
\resizebox{\textwidth}{!}{%
\begin{tikzpicture}[
    font=\small,
    >=Stealth,
    stage/.style={
        rounded corners=2.5mm,
        line width=.75pt,
        minimum height=20mm,
        align=center,
        inner xsep=2.7mm,
        inner ysep=2mm
    },
    link/.style={
        -{Stealth[length=2.3mm,width=1.5mm]},
        line width=1.05pt,
        draw=cInk!72,
        rounded corners=2mm
    },
    qcard/.style={
        rounded corners=2mm,
        line width=.55pt,
        minimum height=22mm,
        text width=33.5mm,
        align=left,
        inner xsep=2.4mm,
        inner ysep=2mm
    },
    branch/.style={
        -{Stealth[length=2mm,width=1.35mm]},
        line width=1pt,
        rounded corners=2mm
    }
]

\definecolor{cInk}{HTML}{253650}
\definecolor{cMuted}{HTML}{65758B}
\definecolor{cBlue}{HTML}{EDF5FD}
\definecolor{cBlueEdge}{HTML}{679DCD}
\definecolor{cPurple}{HTML}{F1EDFA}
\definecolor{cPurpleEdge}{HTML}{9278C9}
\definecolor{cAmber}{HTML}{FFF4DD}
\definecolor{cAmberEdge}{HTML}{CBA55A}
\definecolor{cMint}{HTML}{E4F5EC}
\definecolor{cMintEdge}{HTML}{4EAA82}
\definecolor{cPeach}{HTML}{FCEBE6}
\definecolor{cPeachEdge}{HTML}{CE8876}
\definecolor{cSoft}{HTML}{F7F9FC}
\definecolor{cPinkEdge}{HTML}{C77C9F}

\node[
    stage,
    draw=cBlueEdge,
    fill=cBlue,
    text width=24.5mm
] (data) {
    {\scriptsize\color{cBlueEdge!65!cInk}\bfseries 01 / INPUT}\\[1.3mm]
    {\bfseries Brain MRI slice}\\[1.1mm]
    {\footnotesize BraTS (glioma)}\\[-.1mm]
    {\footnotesize IXI (healthy)}
};

\node[
    stage,
    draw=cPurpleEdge,
    fill=cPurple,
    text width=29mm,
    right=7mm of data
] (vlm) {
    {\scriptsize\color{cPurpleEdge!70!cInk}\bfseries 02 / INFERENCE}\\[1.3mm]
    {\bfseries VLM \footnotesize (8 models)}\\[1.1mm]
    {\footnotesize ``Tumor or gross}\\[-.1mm]
    {\footnotesize abnormality?''}
};

\node[
    stage,
    draw=cAmberEdge,
    fill=cAmber,
    text width=23mm,
    right=7mm of vlm
] (out) {
    {\scriptsize\color{cAmberEdge!70!cInk}\bfseries 03 / OUTPUT}\\[1.3mm]
    {\bfseries Answer $\hat y$}\\[1mm]
    {\bfseries Confidence $c$}
};

\node[
    diamond,
    aspect=1.32,
    draw=cInk!75,
    fill=cSoft,
    line width=.8pt,
    inner sep=1.15mm,
    align=center,
    right=7mm of out
] (gate) {
    {\bfseries $c\ge\tau$}\\[-.6mm]
    {\scriptsize ?}
};

\node[
    stage,
    draw=cMintEdge,
    fill=cMint,
    text width=25mm,
    minimum height=9mm,
    right=11.5mm of gate,
    yshift=8mm
] (act) {
    {\bfseries Act on $\hat y$}
};

\node[
    stage,
    draw=cPeachEdge,
    fill=cPeach,
    text width=25mm,
    minimum height=9mm,
    right=11.5mm of gate,
    yshift=-8mm
] (defer) {
    {\bfseries Defer}\\[.25mm]
    {\footnotesize to clinician}
};

\draw[link] (data) -- (vlm);
\draw[link] (vlm) -- (out);
\draw[link] (out) -- (gate);

\draw[branch,draw=cMintEdge!85!cInk]
    (gate.east) -- ++(3.1mm,0)
    |- node[
        pos=.77,
        above,
        font=\scriptsize\bfseries,
        text=cMintEdge!65!cInk
    ] {YES}
    (act.west);

\draw[branch,draw=cPeachEdge!85!cInk]
    (gate.east) -- ++(3.1mm,0)
    |- node[
        pos=.77,
        below,
        font=\scriptsize\bfseries,
        text=cPeachEdge!65!cInk
    ] {NO}
    (defer.west);

\node[
    qcard,
    draw=cPinkEdge!45,
    fill=cPinkEdge!7,
    anchor=north west
] (q1) at ($(data.south west)+(0mm,-15mm)$) {
    {\bfseries\color{cPinkEdge!80!cInk}Q1\enspace Low-error subset?}\\[1.8mm]
    {\footnotesize At 5\%, 10\%, 20\% error?}\\[1.9mm]
    {\scriptsize\color{cMuted}Sec.~\ref{sec:gate}}
};

\node[
    qcard,
    draw=cPurpleEdge!45,
    fill=cPurpleEdge!7,
    anchor=north west
] (q2) at ($(q1.north east)+(3mm,0)$) {
    {\bfseries\color{cPurpleEdge!80!cInk}Q2 \quad Why not?}\\[1.8mm]
    {\footnotesize Answer-level vs.}\\[-.2mm]
    {\footnotesize within-answer signal}\\[1.4mm]
    {\scriptsize\color{cMuted}Sec.~\ref{sec:mech}}
};

\node[
    qcard,
    draw=cBlueEdge!45,
    fill=cBlueEdge!7,
    anchor=north west
] (q3) at ($(q2.north east)+(3mm,0)$) {
    {\bfseries\color{cBlueEdge!80!cInk}Q3 \quad Case mix?}\\[1.8mm]
    {\footnotesize Does case mix matter?}\\[1.9mm]
    {\scriptsize\color{cMuted}Sec.~\ref{sec:prev}}
};

\node[
    qcard,
    draw=cAmberEdge!55,
    fill=cAmberEdge!9,
    anchor=north west
] (q4) at ($(q3.north east)+(3mm,0)$) {
    {\bfseries\color{cAmberEdge!85!cInk}Q4 \quad Lesion size?}\\[1.8mm]
    {\footnotesize Does confidence track it?}\\[1.9mm]
    {\scriptsize\color{cMuted}Sec.~\ref{sec:size}}
};

\draw[cPinkEdge,line width=2.2pt,rounded corners=.6mm]
    ($(q1.north west)+(2.4mm,-.6mm)$) --
    ($(q1.north east)+(-2.4mm,-.6mm)$);

\draw[cPurpleEdge,line width=2.2pt,rounded corners=.6mm]
    ($(q2.north west)+(2.4mm,-.6mm)$) --
    ($(q2.north east)+(-2.4mm,-.6mm)$);

\draw[cBlueEdge,line width=2.2pt,rounded corners=.6mm]
    ($(q3.north west)+(2.4mm,-.6mm)$) --
    ($(q3.north east)+(-2.4mm,-.6mm)$);

\draw[cAmberEdge,line width=2.2pt,rounded corners=.6mm]
    ($(q4.north west)+(2.4mm,-.6mm)$) --
    ($(q4.north east)+(-2.4mm,-.6mm)$);

\draw[cInk!22,line width=.7pt]
    ($(q1.north west)+(0,7mm)$) --
    ($(q4.north east)+(0,7mm)$);

\node[
    fill=white,
    inner xsep=2mm,
    inner ysep=.4mm,
    font=\scriptsize\bfseries,
    text=cMuted
] at ($(q1.north west)!.5!(q4.north east)+(0,7mm)$) {
    Q1--Q2: BraTS and IXI
    \qquad Q3: BraTS
    \qquad Q4: BraTS+
};

\end{tikzpicture}%
}

\caption{Confidence-gated MRI assessment and evaluation questions.
A vision--language model predicts an answer $\hat y$ and reports
stated confidence $c$ for a brain-MRI slice. The gate accepts the
prediction when $c\ge\tau$ and otherwise defers the case to a clinician.
We evaluate confidence-gated selection under specified empirical
error budgets (Q1) and answer-conditioned confidence (Q2) separately
on BraTS and IXI. We examine case-mix sensitivity by reweighting
within BraTS (Q3), and lesion-size dependence on BraTS-positive
slices (Q4).}
\label{fig:overview}
\end{figure*}
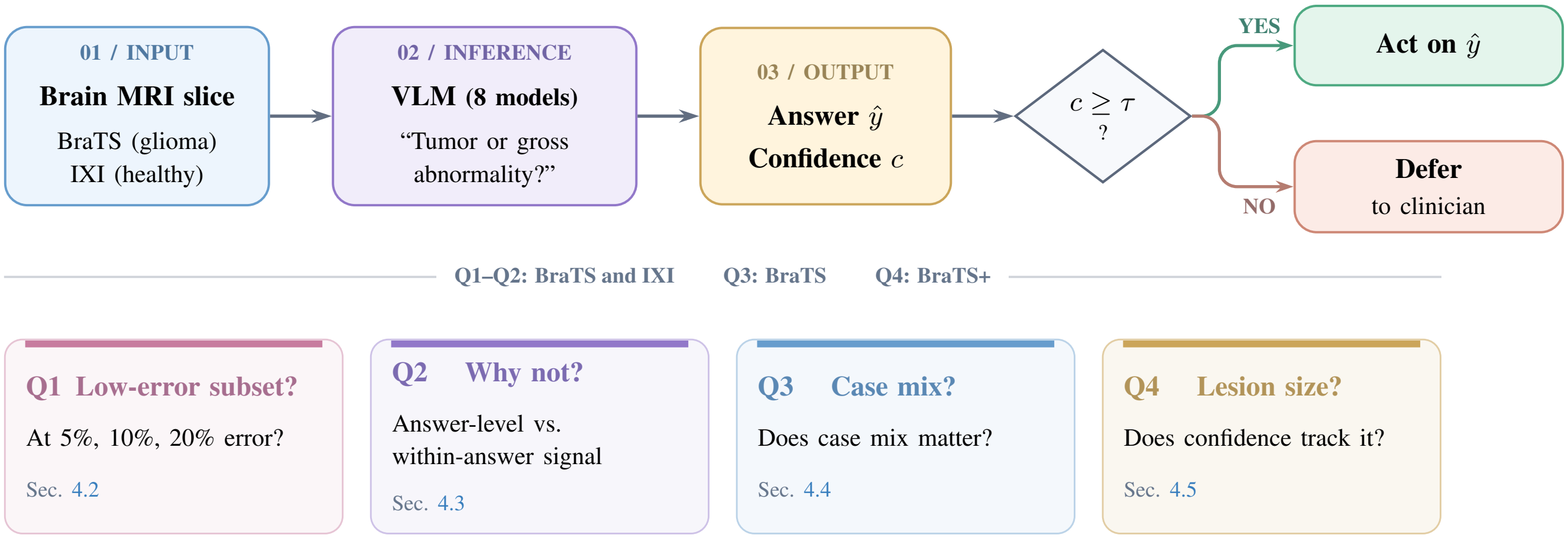

\section{Introduction}
Foundation vision--language models (VLMs) can read a medical image and answer a question about it in plain language. This makes them tempting parts of clinical decision support systems (CDSS), for example to pre-read studies or flag findings. Any such system needs a rule for when to act on the model's answer and when to hand the case to a clinician. The standard design is a deferral gate: act when a reliability score clears a threshold, and refer the rest. This is the classical reject option~\cite{chow1970optimum}, and clinical AI uses it to split work between models and human readers~\cite{leibig2022combining,dvijotham2023enhancing}.

For a chat-style VLM, the most available reliability score is the confidence the model states with its answer. It needs no extra inference pass and no access to model internals, and it is exactly what a clinician sees on screen. It is also a claim the model makes about itself: that one answer deserves more trust than another. For a CDSS, that claim has a direct test: does thresholding it leave a set of answers reliable enough to act on?

Most evidence about stated confidence comes from calibration, and it is not encouraging: language models and medical visual question answering (VQA) systems tend to be overconfident~\cite{xiong2024can,byun2026overconfidence}, although eliciting it carefully can help~\cite{tian2023just}. Calibration, however, does not answer the question a gate asks. One summary score can hide distinct failure modes~\cite{maier2024metrics,sabbaghziarani2026label}. A miscalibrated score can still rank cases well, and a well-calibrated one can fail exactly where the gate operates. Recent work therefore judges confidence estimators for medical VLMs by how much work they can safely automate~\cite{khanmohammadi2026calibrated}. Two questions remain open for image-level clinical decisions. First, does a confidence gate keep its meaning when the population changes, for instance from a neuro-oncology clinic to people without disease? Second, when it fails, what in the confidence signal makes it fail?

We study both questions on brain MRI, where a reference can be derived for every slice. Eight open-weight VLMs, including two reasoning models, see 2-D slices from post-treatment glioma patients and healthy volunteers and answer several questions about each one. We build the deferral analysis on the question a CDSS would act on: whether the slice shows a tumor or gross abnormality. We then evaluate stated confidence as a gate, separately for each population (Fig.~\ref{fig:overview}). Our contributions are:
\begin{itemize}
\item We evaluate stated confidence as a deferral gate on each population. On glioma patients, no threshold meets a 5\% error budget, and only one model passes a 10\% budget on more than 3\% of slices (Sections~\ref{sec:gate} and~\ref{sec:checks}).
\item We decompose the covariance between confidence and correctness into a part tied to the answer and a part within each answer. This explains how the correlation can change sign between populations; separately, a rule that ignores confidence keeps more slices than the gate for two models (Sections~\ref{sec:mech} and~\ref{sec:prev}).
\item We find that mean stated confidence changes little with lesion area, even where sensitivity nearly doubles (Section~\ref{sec:size}).
\end{itemize}

\section{Related Work}
Our work draws on three lines of research: deciding when to defer, estimating the confidence of (vision--)language models, and evaluating medical VLMs.

\textbf{Deferral and selective prediction.} Chow's reject option~\cite{chow1970optimum} formalized the trade between error and abstention. Selective classification~\cite{geifman2017selective} and risk--coverage analysis~\cite{geifman2019bias} made it standard for deep networks, and learning-to-defer methods train the classifier and the deferral rule jointly~\cite{madras2018predict,mozannar2020consistent}. In clinical imaging, decision referral lets a mammography model handle confident cases and pass the rest to radiologists~\cite{leibig2022combining}. Complementarity-driven deferral learns when to trust a model over a clinician~\cite{dvijotham2023enhancing}. These methods fit their score on data that resemble the deployment data. When the class balance shifts, decisions based on posteriors shift with it~\cite{saerens2002adjusting,lipton2018detecting}. Section~\ref{sec:mech} shows a specific form of this problem for stated confidence.

\textbf{Confidence of language and vision--language models.} Language models can report their own uncertainty~\cite{kadavath2022language}, and for RLHF-tuned models verbalized confidence can be better calibrated than token probabilities~\cite{tian2023just}. It is still often overconfident and only weakly discriminative~\cite{xiong2024can}, and it can track a model's commitment to an answer more than the answer's correctness~\cite{kumaran2026reported}. Medical VQA shows the same pattern~\cite{byun2026overconfidence}. Khanmohammadi et al.~\cite{khanmohammadi2026calibrated} therefore compare estimators by what they can safely automate, with held-out and distribution-free operating points. Across five VLMs and three medical VQA datasets, nine estimators, from self-reports and training-free logit scores to trained probes, differ most in the high-confidence region a clinician would act on. Dhinagar et al.~\cite{dhinagar2026calm} fine-tune a 3-D VLM for brain-MRI classification and abstain when its calibrated token probability is low. These studies compare estimators. We ask a complementary question about the one signal every chat interface exposes: why does stated confidence fail as a gate, and how does its failure depend on the population?

\textbf{Medical VLMs on brain MRI.} Brain MRI is a convenient test bed, because slice-level labels can be derived from public annotations. Audits there report near-complete answer coverage but poorly calibrated stated confidence, with mean confidence of 0.82--0.97 on wrong answers~\cite{sabbaghziarani2026confident}. Medical fine-tuning can even degrade visual discrimination~\cite{zhu2026lost}. More broadly, post-hoc explanations cannot certify an individual decision in health care~\cite{ghassemi2021false}. A model's statement about its own reliability is different, because it can be checked against outcomes. This paper makes that check, separately for each population.

\section{Study Design}
Figure~\ref{fig:overview} summarizes the study: a model answers a question about one slice and states its confidence, and a gate decides whether to act on the answer or defer. This section describes the questions and data (Sec.~\ref{sec:data}), the models (Sec.~\ref{sec:models}), the gate metrics (Sec.~\ref{sec:metrics}) and how we trace a gate's failure (Sec.~\ref{sec:decomp}).

\begin{figure*}[t]
    \centering
    \includegraphics[width=\textwidth]{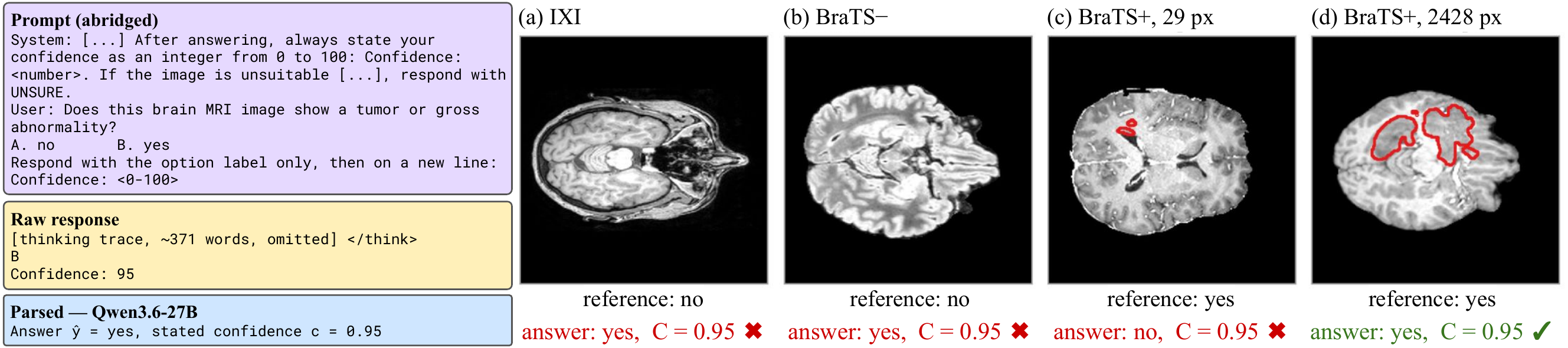}
    \caption{Example exchange and MRI slices.
    The prompt, raw response, and parsed answer are shown on the left.
    Four example slices are shown on the right with their reference
    labels, model answers, and stated confidence.
    Red outlines indicate the reference masks.
    Panels (a) and (b) show false positives, panel (c) shows a
    missed lesion, and panel (d) shows a correctly detected lesion.
    Crosses indicate incorrect predictions; the check mark
    indicates a correct prediction.}
    \label{fig:examples}
\end{figure*}

\subsection{Questions, data and reference}
\label{sec:data}
Each model sees one 2-D slice with a question and must answer with an option and a confidence from 0 to 100. Following the brain-MRI exam of~\cite{sabbaghziarani2026confident}, a slice can be asked about its MRI sequence, its plane, whether it is a brain MRI, whether it shows a tumor or gross abnormality, and on which side a lesion lies, each as multiple choice or as an open question. Our deferral analysis uses the tumor question, as multiple choice with neutral wording, because it is the one a CDSS would act on: whether to flag a slice for a reader.

Slices are taken at 40\%, 50\% and 60\% depth in axial, coronal and sagittal planes, with robust percentile intensity clipping and 8-bit scaling. Post-treatment glioma patients come from BraTS 2024~\cite{deverdier2024brats,menze2015brats,bakas2017advancing} and healthy volunteers from IXI~\cite{ixi}. A BraTS slice is reference-positive if its segmentation mask covers at least 25 pixels on that slice (1\,px $=$ 1\,mm$^2$). Any annotated compartment counts: enhancing tissue, non-enhancing tumor core, surrounding FLAIR hyperintensity (which includes radiation-related change and gliosis) or resection cavity~\cite{deverdier2024brats}. The reference therefore marks annotated tumor or treatment-related change. This overlaps with the question's ``gross abnormality'' but is not identical to it, and it comes from a multimodal annotation rather than a reading of each sequence. IXI slices are treated as negatives: the volunteers are healthy, but no expert read each displayed slice.

Table~\ref{tab:strata} lists the three resulting strata. BraTS+ and BraTS$-$ are the mask-positive and mask-negative slices of the glioma patients. Together we call them \emph{BraTS}: one source with both classes and 70\% positives. IXI has no positives. We exclude 74 BraTS slices without a usable mask and 70 whose mask covers only 1--24\,px, too little to call either way. BraTS images are distributed skull-stripped and co-registered to a common atlas, but IXI images are not. IXI slices therefore keep the skull, the scalp and, at inferior levels, the orbits and skull base (Fig.~\ref{fig:examples}a). All data are public and de-identified.

Figure~\ref{fig:examples} shows one exchange end to end. The model receives the slice, the question and the two options in random order. A reasoning model first writes a thinking trace. Every model then returns an option and a confidence, which we parse into its answer and stated confidence. The example slices were drawn at random, with a fixed seed, from four outcome types: two kinds of false positive, a missed lesion and a found lesion.

\begin{table}[t]
\caption{The three strata of the tumor decision.}
\label{tab:strata}
\centering
\footnotesize
\begin{tabular}{@{}llrrl@{}}
\toprule
Stratum & Source & Subjects & Slices & Reference \\
\midrule
BraTS+ & BraTS 2024 & \multirow{2}{*}{150} & 840 & positive \\
BraTS$-$ & BraTS 2024 & & 366 & negative \\
IXI & IXI & 100 & 2,682 & negative \\
\bottomrule
\end{tabular}

\vspace{2pt}
\parbox{\linewidth}{\scriptsize BraTS+ and BraTS$-$ come from the same pool of 150 patients. Sequences: T1 and T2 (both sources), T1ce and FLAIR (BraTS only), PD (IXI only).}
\end{table}

\subsection{Models and inference}
\label{sec:models}
We evaluate six instruction-tuned models: InternVL2.5-8B~\cite{chen2024internvl25}, Qwen2.5-VL-3B and -7B~\cite{bai2025qwen25vl}, Gemma-3-4B~\cite{gemma2025gemma3}, its medical fine-tune MedGemma-4B~\cite{sellergren2025medgemma}, and Gemma-4-12B~\cite{gemma2026gemma4}. We add two reasoning models that write a thinking trace before answering: GLM-4.6V-Flash (9B)~\cite{glm2025glm41v,zai2025glm46vflash} and Qwen3.6-27B~\cite{qwen2026qwen36}. All run locally in bfloat16 with Hugging Face Transformers and greedy decoding. They may generate up to 512 new tokens, or 2,048 for GLM-4.6V-Flash and 8,192 for Qwen3.6-27B so that the trace can finish. The yes/no order is randomized per slice and logged, and the chosen letter is mapped back through it. For reasoning models we parse only the text after the final \texttt{</think>}. Confidence is the first integer from 0 to 100 after ``Confidence'', rescaled to $[0,1]$. A response with no parseable option, an UNSURE, or a trace cut off at the token limit counts as deferred. Prompts and parsing follow~\cite{sabbaghziarani2026confident}, which audited the six instruction-tuned models on all five questions; the two reasoning models were run for this study.

\subsection{Notation and metrics}
\label{sec:metrics}
For each slice, $y\in\{0,1\}$ is the reference, $\yhat$ the model's answer, $z=\mathbf{1}[\yhat=y]$ whether it is right, and $c\in[0,1]$ the stated confidence. Besides sensitivity, specificity and balanced accuracy~\cite{brodersen2010balanced}, we use four quantities. The share of ``yes'' answers is $q$. The predictive values PPV and NPV say how often a ``yes'' or a ``no'' answer is right. The correlation between $c$ and $z$ is $\rho$. The \emph{case mix} $\pi$ is the share of reference-positive slices; within BraTS it reflects slice sampling, not how many patients have a tumor. The \emph{outcome-level confidence} is the mean $c$ over true positives, false positives, true negatives or false negatives. All rates use answered slices.

A deferral gate with threshold $\tau$ accepts the answers with $c\ge\tau$ and defers the rest. On $n$ slices, its \emph{coverage} $\phi(\tau)$ and \emph{selective risk} $R(\tau)$ are
\begin{equation}
\phi(\tau)=\frac{1}{n}\sum_{i=1}^{n}\mathbf{1}[c_i\ge\tau],\qquad
R(\tau)=\frac{\sum_{i}\mathbf{1}[c_i\ge\tau]\,(1-z_i)}{\sum_{i}\mathbf{1}[c_i\ge\tau]}.
\label{eq:cov}
\end{equation}
Stated confidences take few distinct values, and most models put hundreds of slices on a single value such as 0.90. A threshold cannot split such a tie, so we try every value a model uses and report
\begin{equation}
C_\alpha=\max_{\tau}\,\{\phi(\tau): R(\tau)\le\alpha\},\qquad \alpha\in\{0.05,0.10,0.20\},
\label{eq:calpha}
\end{equation}
the largest share of slices that any threshold can accept with an error rate of at most $\alpha$, or 0 if none can. It is an in-sample version of the yield at an error budget used in~\cite{khanmohammadi2026calibrated}, which chooses thresholds on held-out data. Because the threshold is chosen after seeing the outcomes, a zero is strong negative evidence, while a positive value is the best coverage observed in this sample, not a guarantee for new data. Deferred slices count in $n$ but are never accepted, so coverage measures how much of this slice task could be automated, not clinical workload. The targets are stress tests, not clinical thresholds. Intervals are 95\% subject-clustered bootstrap intervals with 1,000 resamples~\cite{efron1993bootstrap}.

\subsection{Where a correlation comes from}
\label{sec:decomp}
A gate works only if confidence separates right from wrong answers. Confidence can also correlate with correctness for a simpler reason. A model may state one answer more confidently than the other, and on a given population one answer may be right more often. To tell these apart, let $\mu_1$ and $\mu_0$ be the mean confidence on ``yes'' and ``no'' answers, and call the answer with the larger mean the model's \emph{more-confident answer}. Below, $\mathrm{Cov}(c,z)$ is the covariance between confidence and correctness over a population, $\mathrm{Cov}(c,z\mid\yhat=1)$ is the same covariance among the ``yes'' answers only, and $\mathbb{E}$ denotes a mean. The law of total covariance, applied to the answer, then gives the split we use.

\begin{lemma}[Answer decomposition]
\label{prop:answer}
For any confidence signal $c$, with the share of ``yes'' answers $q$, the predictive values PPV and NPV, and the mean confidences $\mu_1$ and $\mu_0$ all measured on the same population,
\begin{equation}
\begin{aligned}
\mathrm{Cov}(c,z)={}&q(1-q)\,(\mu_1-\mu_0)\,(\mathrm{PPV}-\mathrm{NPV})\\
&+q\,\mathrm{Cov}(c,z\mid\yhat=1)\\
&+(1-q)\,\mathrm{Cov}(c,z\mid\yhat=0).
\end{aligned}
\label{eq:cov-identity}
\end{equation}
We call the first term the \emph{answer term} and the other two the \emph{within-answer terms}.
\end{lemma}
\emph{Proof.} By the law of total covariance, $\mathrm{Cov}(c,z)=\mathbb{E}[\mathrm{Cov}(c,z\mid\yhat)]+\mathrm{Cov}(\mathbb{E}[c\mid\yhat],\mathbb{E}[z\mid\yhat])$. The first part gives the within-answer terms. For a binary $\yhat$, the second equals the answer term, since $\mathbb{E}[z\mid\yhat=1]=\mathrm{PPV}$ and $\mathbb{E}[z\mid\yhat=0]=\mathrm{NPV}$. \hfill$\blacksquare$

The answer term is positive when the more-confident answer is also the more reliable one on the population at hand. It depends on the population, not on what the model knows about its errors. The within-answer terms are positive when, within the same answer, correct answers receive higher mean confidence than wrong ones. Such differences can support selection beyond the answer itself; whether they do is judged by the error of the accepted set. Two consequences guide the analysis:

\begin{corollary}
\label{cor:answer}
(a) If every slice has the same true label, as on IXI, the within-answer terms are zero for any confidence signal, so the sign of $\rho$ only shows which answer is stated more confidently. (b) If confidence depends only on the answer, a gate can accept only a whole answer class or every answer.
\end{corollary}

\section{Results}
\subsection{The tumor call is weak, and confidence does not show it}
\label{sec:call}

\begin{table}[t]
\caption{The tumor call by stratum.}
\label{tab:call}
\centering
\footnotesize
\setlength{\tabcolsep}{3pt}
\begin{tabular}{@{}llccccc@{}}
\toprule
Model & Type & Sens. & Spec. & Spec. & BA & Conf. \\
 & & BraTS+ & BraTS$-$ & IXI & BraTS & \\
\midrule
InternVL2.5-8B  & G & .423 & .615 & .957 & .519 & .88--.90 \\
Qwen2.5-VL-3B   & G & .198 & .872 & .707 & .535 & .67--.73 \\
Qwen2.5-VL-7B   & G & .042 & .984 & .837 & .513 & .67--.70 \\
Gemma-3-4B      & G & .619 & .473 & .274 & .546 & .87--.89 \\
MedGemma-4B     & M & .424 & .814 & .727 & .619 & .95 \\
Gemma-4-12B     & G & .616 & .702 & .525 & .659 & .92--.94 \\
GLM-4.6V-Flash  & R & .637 & .573 & .626 & .605 & .80--.81 \\
Qwen3.6-27B     & R & .690 & .601 & .690 & .646 & .95--.96 \\
\bottomrule
\end{tabular}

\vspace{2pt}
\parbox{\linewidth}{\scriptsize Type: G general, M medical, R reasoning. BA: balanced accuracy. Conf.: mean stated confidence (range over strata).}
\end{table}

No model makes this call reliably (Table~\ref{tab:call}). The models differ mostly in how readily they say ``yes''. At one end, Qwen2.5-VL-7B almost never does and misses nearly every lesion; at the other, Gemma-3-4B calls most healthy slices positive. Most models trade sensitivity against specificity instead of improving both, and none reaches a balanced accuracy of 0.70 on BraTS. The two reasoning models are the most balanced across the three strata, but not clearly more accurate. Their subject-clustered 95\% intervals stay within $\pm0.05$ of these rates.

Stated confidence does not reflect this weakness. Every model reports high confidence on every stratum, including MedGemma-4B, which states 0.95 while missing more than half of the lesions. The models also disagree with each other. On average, only 3.6 of the eight call a BraTS+ slice correctly, and all eight agree with the mask on fewer than 1\% of these slices.

The six instruction-tuned models answer essentially every slice, and the two reasoning models more than 99\%. The token budget mainly affected coverage. At 2,048 tokens, Qwen3.6-27B answered only 66.8\% of slices; on the slices answered under both budgets, its sensitivity changed by less than 0.01.

\subsection{No threshold meets a strict error target on BraTS}
\label{sec:gate}

\begin{table}[t]
\caption{Stated confidence as a deferral gate on BraTS.}
\label{tab:gate}
\centering
\footnotesize
\begin{tabular}{@{}lcccc@{}}
\toprule
Model & Acc. & $C_{0.05}$ & $C_{0.10}$ & $C_{0.20}$ \\
\midrule
InternVL2.5-8B  & .481 & 0 & 0 & 0 \\
Qwen2.5-VL-3B   & .402 & 0 & 0 & 0 \\
Qwen2.5-VL-7B   & .328 & 0 & 0 & .034 (41) \\
Gemma-3-4B      & .575 & 0 & 0 & 0 \\
MedGemma-4B     & .542 & 0 & 0 & 0 \\
Gemma-4-12B     & .642 & 0 & .244 (294) & .483 (582) \\
GLM-4.6V-Flash  & .617 & 0 & .025 (30) & .025 (30) \\
Qwen3.6-27B     & .663 & 0 & 0 & .578 (697) \\
\bottomrule
\end{tabular}

\vspace{2pt}
\parbox{\linewidth}{\scriptsize Acc.: accuracy. $C_\alpha$ (Eq.~\ref{eq:calpha}): largest share of slices a threshold can accept with at most $\alpha$ errors; accepted slices in parentheses.}
\end{table}

Searching every threshold, no model's confidence selects any set of BraTS slices that meets a 5\% empirical error target (Table~\ref{tab:gate}). At 10\%, the picture depends on the model: Gemma-4-12B keeps about a quarter of the slices, while every other model keeps at most 2.5\%. Only at 20\% do several models keep something. For Gemma-4-12B and Qwen2.5-VL-7B at 20\%, Section~\ref{sec:mech} shows that accepting whole answer classes keeps as much or more, so their coverage need not reflect knowledge of their own errors.\IfFileExists{figures/risk_coverage.pdf}{ Figure~\ref{fig:rc} shows the full curves behind Table~\ref{tab:gate}; no curve enters the 5\% band.}{}

\IfFileExists{figures/risk_coverage.pdf}{%
\begin{figure}[t]
\centering
\includegraphics[width=\columnwidth]{figures/risk_coverage.pdf}
\caption{Risk--coverage curves on BraTS (Eq.~\ref{eq:cov}), one step per confidence value. Dotted lines: 5\%, 10\% and 20\% targets. Hollow markers: answer-only rule (Section~\ref{sec:mech}).}
\label{fig:rc}
\end{figure}}{}

Healthy volunteers tell a different story. On IXI every slice is negative, so the gate's error is simply the share of ``yes'' answers. The two most specific models, InternVL2.5-8B and Qwen2.5-VL-7B, meet a 5\% and a 20\% target there without deferring anything, yet both are among the least sensitive on BraTS. A gate validated on healthy brains would thus look safest for the models that miss the most tumors.

Calibration error does not decide this. $C_\alpha$ depends only on how confidence orders the slices, so any strictly increasing recalibration, such as temperature scaling, leaves it unchanged.

\begin{table*}[t]
\caption{The confidence gate against a rule that ignores confidence, on BraTS. $^\dagger$Inferred from the sign of $\rho$ on IXI, assuming the same answer on BraTS. Like $C_\alpha$, the answer-only rule is chosen in-sample: it is the best of accepting all ``yes'', all ``no'' or all answers.}
\label{tab:ceiling}
\centering
\small
\setlength{\tabcolsep}{7pt}
\begin{tabular}{@{}lccccc@{}}
\toprule
 & More-confident & Share right & Share of slices & \multicolumn{2}{c}{Coverage at a 20\% target} \\
\cmidrule(l){5-6}
Model & answer & (PPV or NPV) & given that answer & confidence gate & answer-only rule \\
\midrule
InternVL2.5-8B  & no            & .317 & .589 & 0 & 0 \\
Qwen2.5-VL-3B   & yes$^\dagger$ & .780 & .177 & 0 & 0 \\
Qwen2.5-VL-7B   & yes$^\dagger$ & .854 & .034 & .034 & .034 \\
Gemma-3-4B      & no$^\dagger$  & .351 & .409 & 0 & 0 \\
MedGemma-4B     & no$^\dagger$  & .381 & .648 & 0 & .352 \\
Gemma-4-12B     & yes           & .826 & .519 & .483 & .519 \\
GLM-4.6V-Flash  & yes           & .774 & .573 & .025 & 0 \\
Qwen3.6-27B     & yes           & .798 & .598 & .578 & 0 \\
\bottomrule
\end{tabular}
\end{table*}

\subsection{Where the association comes from}
\label{sec:mech}

\begin{figure*}[t]
\centering
\includegraphics[width=\textwidth]{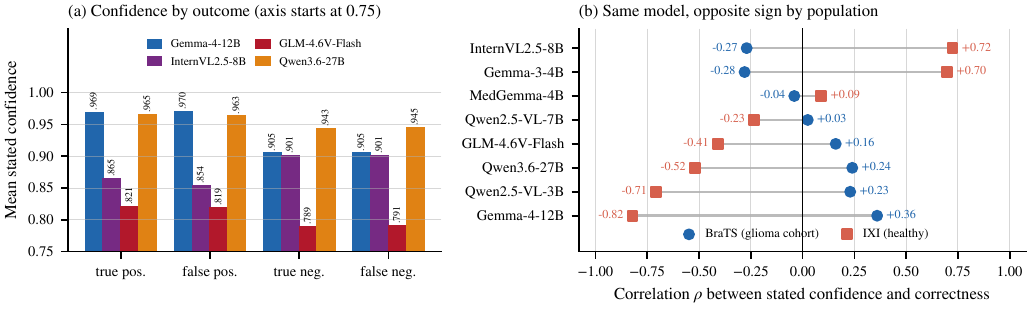}
\caption{Confidence differs more between answers than between right and wrong answers. (a) Outcome-level confidence (mean stated confidence of true and false positives and negatives), pooled over all labeled slices: true and false positives receive nearly the same confidence, and so do true and false negatives. (b) Correlation $\rho$ between stated confidence and correctness on BraTS (circles) and IXI (squares). On IXI its sign only records which answer a model states more confidently (Corollary~\ref{cor:answer}a); on BraTS it has the opposite sign for all eight models.}
\label{fig:mech}
\end{figure*}

For the four models with outcome-level confidences, pooled over both populations, confidence differs far more between answers than between right and wrong answers to the same question (Fig.~\ref{fig:mech}a). Gemma-4-12B, for example, states the same confidence on its true and false positives (0.969 and 0.970) and on its true and false negatives (0.905). To see where the correlation itself comes from, we apply the decomposition of Section~\ref{sec:decomp}.

\emph{Healthy volunteers.} On IXI, the sign of $\rho$ only records which answer a model states more confidently (Corollary~\ref{cor:answer}a). Its practical meaning is what matters: for a model that is more confident when it says ``yes'', its wrong ``yes'' answers on IXI receive higher mean confidence than its correct ``no'' answers. Gemma-4-12B is such a model ($\rho=-0.821$ on IXI).

\emph{Glioma patients.} On BraTS, every model's ``yes'' answers are right more often than its ``no'' answers, so if the more-confident answer is the same as on IXI, the answer term takes the opposite sign. The point estimates of $\rho$ do flip sign for all eight models (Fig.~\ref{fig:mech}b), most weakly where they are near zero (MedGemma-4B, Qwen2.5-VL-7B). For three of the four models with outcome-level means, an approximate decomposition is consistent with this reading; for InternVL2.5-8B, a small within-answer signal pulls the other way. We treat this as exploratory, because some of these means are pooled over both populations. The decomposition also explains the correlation, not the gate: a small within-answer term can still hide a useful high-confidence subset, so gate performance comes from the threshold search. Pooled over both populations, Gemma-4-12B's correlation is $-0.461$: one number that hides two opposite relationships.

\emph{What a gate can do.} On BraTS, every model's ``yes'' answers are wrong at least 14\% of the time and its ``no'' answers at least 54\%. Accepting a whole answer class can therefore never meet a 5\% target; only confidence differences within an answer could, and no model has enough of them. Table~\ref{tab:ceiling} compares the gate with a rule that ignores confidence. At a 20\% target, the rule keeps more slices than the gate for Gemma-4-12B and MedGemma-4B. MedGemma-4B states ``no'' more confidently on average (inferred from IXI and assumed to hold on BraTS), which favors its less reliable answer class. Beyond selecting whole answer classes, confidence helps only where it separates right from wrong within an answer. The clearest cases are Gemma-4-12B at 10\% and Qwen3.6-27B at 20\%, whose ``yes'' answers miss that target by only a quarter of a point. In the accepted sets at 20\%, every error of Gemma-4-12B and Qwen2.5-VL-7B is a false positive.\IfFileExists{figures/risk_coverage.pdf}{ In Figure~\ref{fig:rc}, the answer-only points of Gemma-4-12B and MedGemma-4B lie below their confidence curves.}{}

\begin{table*}[t]
\caption{Large-lesion, distant-negative subset of BraTS: 480 positive and 140 negative slices, reweighted to the natural case mix of 0.70. \emph{Specificity needed}: the specificity at which all of a model's ``yes'' answers together would have at most 5\% error, $1-\frac{\alpha}{1-\alpha}\cdot\frac{\pi}{1-\pi}\cdot\text{sensitivity}$ with $\alpha=0.05$, $\pi=0.70$. $^\dagger$Computed from the subset's sensitivity and specificity.}
\label{tab:clear}
\centering
\small
\setlength{\tabcolsep}{6pt}
\begin{tabular}{@{}lccccccc@{}}
\toprule
 & \multicolumn{2}{c}{Measured on the subset} & Specificity needed & \multicolumn{3}{c}{Coverage $C_\alpha$, all BraTS $\rightarrow$ subset} & Answer-only \\
\cmidrule(lr){2-3}\cmidrule(lr){5-7}
Model & Sensitivity & Specificity & for a 5\% target & $\alpha=0.05$ & $\alpha=0.10$ & $\alpha=0.20$ & $C_{0.20}$, subset \\
\midrule
Gemma-4-12B     & 0.68 & 0.70 & 0.92 & $0 \rightarrow 0$ & $0.244 \rightarrow 0.31$ & $0.483 \rightarrow 0.52$ & 0.57$^\dagger$ \\
GLM-4.6V-Flash  & 0.71 & 0.57 & 0.91 & $0 \rightarrow 0$ & $0.025 \rightarrow 0.03$ & $0.025 \rightarrow 0.56$ & 0$^\dagger$ \\
Qwen3.6-27B     & 0.75 & 0.60 & 0.91 & $0 \rightarrow 0$ & $0 \rightarrow 0$ & $0.578 \rightarrow 0.59$ & 0.65$^\dagger$ \\
MedGemma-4B     & 0.49 & 0.81 & 0.94 & $0 \rightarrow 0$ & $0 \rightarrow 0$ & $0 \rightarrow 0$ & 0.40$^\dagger$ \\
\bottomrule
\end{tabular}
\end{table*}

\subsection{Changing the case mix within one source}
\label{sec:prev}

\IfFileExists{figures/risk_coverage.pdf}{}{%
\begin{figure}[t]
\centering
\includegraphics[width=\columnwidth]{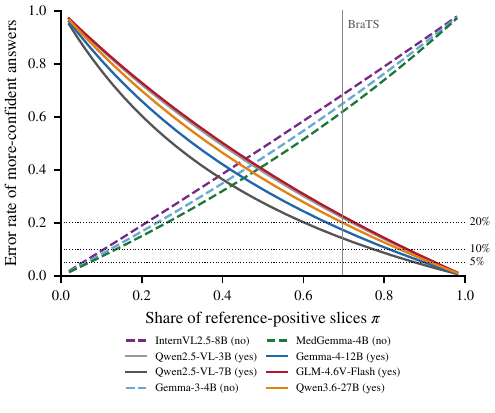}
\caption{Error rate of each model's more-confident answers, taken as a whole class rather than through a confidence threshold, as the case mix $\pi$ changes (Bayes' rule on Table~\ref{tab:call}). Solid: ``yes''; dashed: ``no''. Dotted: 5\%, 10\%, 20\% targets; gray: natural BraTS mix.}
\label{fig:prev}
\end{figure}}

The sign flip in Section~\ref{sec:mech} compares two sources that also differ in scanners, sequences and skull-stripping. To isolate case mix, we reweight BraTS so that a share $\pi$ between 0.10 and 0.90 of slices is reference-positive. Each model's sensitivity and specificity stay fixed.\IfFileExists{figures/risk_coverage.pdf}{}{ Figure~\ref{fig:prev} shows the result for each model's more-confident answers.} Their error rate moves steeply with $\pi$. Gemma-4-12B's ``yes'' answers go from 17\% wrong at the natural mix to 81\% wrong at $\pi=0.10$, while models whose more-confident answer is ``no'' move the other way. This explains why the reliability of a model's favored answers, and with it what a gate can keep, changes with case mix; it does not test a threshold carried from one case mix to another. This follows from Bayes' rule under label shift~\cite{saerens2002adjusting,lipton2018detecting}; it does not show that the BraTS--IXI difference is due to case mix alone. The answer term changes sign where PPV equals NPV, provided the more-confident answer stays the same; for the four models with outcome-level means, this happens at $\pi$ of about 0.49--0.51. Whether the total correlation also changes sign depends on the size and direction of the within-answer terms.

\subsection{Mean confidence barely changes with lesion area}
\label{sec:size}

\begin{figure}[t]
\centering
\includegraphics[width=\columnwidth]{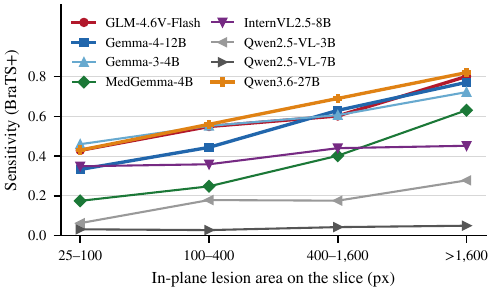}
\caption{Sensitivity on BraTS+ by the in-plane area of the annotated lesion on the slice (1\,px $=$ 1\,mm$^2$). Mean stated confidence over the same slices changes little across bins (Section~\ref{sec:size}).}
\label{fig:size}
\end{figure}

Sensitivity rises steeply with lesion area for most models (Fig.~\ref{fig:size}); on the smallest lesions (0.25--1\,cm$^2$) it is only 0.03--0.46. Mean stated confidence barely follows. From the smallest to the largest bin it moves from 0.80 to 0.82 for GLM-4.6V-Flash, from 0.93 to 0.95 for Gemma-4-12B, and from 0.95 to 0.96 for Qwen3.6-27B, whose sensitivity nearly doubles; for the other five models it stays near a model-specific value between 0.67 and 0.95. A clinician reading the stated confidence would get little warning on the slices where a miss is most likely. These are means, so they do not rule out ranking within a bin. Because every slice here is positive, part of any change in mean confidence also comes from the shift from ``no'' to ``yes'' answers.

\subsection{Robustness checks}
\label{sec:checks}
\emph{Positivity threshold.} Raising the mask threshold from 25 to 100\,px changes sensitivity by at most 0.021 and specificity by at most 0.005 for the four models we checked. Table~\ref{tab:call} therefore does not hinge on the 25\,px choice.

\emph{Large-lesion, distant-negative subset.} Is the gate failing only because some references are debatable? As a partial check, we keep BraTS+ slices with at least 400\,px of annotated lesion on T2, FLAIR or T1ce, and BraTS$-$ slices with no mask voxel within 10\,mm. We then reweight them to the natural case mix (Table~\ref{tab:clear}). This restricts the images; it does not re-read them. The strict result does not move: the threshold search again keeps nothing at a 5\% target for all four models. Accepting all ``yes'' answers would also fail it, since at these sensitivities it would need a specificity of 0.91--0.94, against 0.57--0.81 observed. This calculation concerns the whole answer class; the threshold search separately shows that no smaller confidence-selected set qualifies. The looser targets do move, sometimes a lot: GLM-4.6V-Flash rises from 2.5\% to 56\% at 20\%, because its ``yes'' answers now sit just above the target. The answer-only rule gains too, and keeps more slices than the gate for three of the four models.

\section{Discussion}
\textbf{What this means for a CDSS.} Raw stated confidence should not be the switch between acting and deferring, though it is not useless: for Gemma-4-12B it keeps about a quarter of BraTS slices at a 10\% target, so the answer depends on the model and on the target. Neither reasoning model changes this: Qwen3.6-27B has the highest accuracy on BraTS (0.663), yet neither it nor GLM-4.6V-Flash keeps a slice at a 5\% target. Lemma~\ref{prop:answer} suggests two cheap checks before deployment. First, tabulate confidence by outcome and by population, and compute~\eqref{eq:cov-identity}. If the answer term dominates, the gate inherits the population, and the answer-only rule is the baseline to beat. Second, validate the gate at a stated error target on each population it will serve, with thresholds chosen on separate subjects. The share of positive cases differs between referral clinics, surveillance imaging and incidental findings, so the same threshold could pass a model's best answers in one setting and its worst in another; how such differences map onto slice-level case mix is untested.

\textbf{How this fits other evidence.} Across medical VQA datasets, trained internal probes and training-free logit scores often supported safer automation than inference-only self-reports, with strong model and domain dependence~\cite{khanmohammadi2026calibrated}; and calibrated token probabilities of a fine-tuned model support useful selective prediction~\cite{dhinagar2026calm}. Our results agree with both: the weak link is the stated confidence of untuned models, not uncertainty estimation for VLMs in general. Kumaran~\cite{kumaran2026reported} reaches a related conclusion for language models: verbal confidence follows the model's commitment, while token log-probabilities follow correctness. The next test is to run answer-token likelihood and sampling agreement~\cite{wang2023selfconsistency} through the same per-population gate.

\textbf{Limitations.} The reference comes from masks, not from a radiologist's read of each slice. A compartment delineated on the multimodal study need not be visible on the single sequence shown. Mask-negative slices from treated patients may also show change outside the annotation. Both can turn a reasonable answer into a scored error, and $C_{0.05}$ is the figure most exposed to them. The large-lesion, distant-negative subset leaves it at zero, but it restricts the images rather than re-reading them; a blinded expert read of the displayed images and results by sequence are the next checks. BraTS and IXI differ in scanners, sequences and skull-stripping, so a model can tell them apart from the image surround; the reweighting within BraTS avoids this confound. $C_\alpha$ is chosen in-sample and weights both error types equally. We report no intervals for it, for the decomposition or for the lesion bins. The decomposition covers four models and uses outcome-level means that are partly pooled over populations, and lesion area is only one aspect of difficulty. Finally, we study 2-D slices of one binary decision, with greedy decoding and stated confidence only, and the public datasets may appear in pretraining corpora.

\textbf{Data, ethics and release.} BraTS 2024 and IXI are public, de-identified datasets used under their terms; we collected no new data and trained no model. The main risk is over-trust: our negative result on one decision is not a verdict on any model, and we make no diagnostic claim. Per-slice outputs and code will be released with the camera-ready version.

\section{Conclusion}
We asked whether a VLM's stated confidence can decide when its brain-MRI tumor call should be trusted. At a strict 5\% error target it cannot, on this sample and against the mask-derived reference: no confidence threshold selects any set of slices that meets the target, also on a subset restricted to large lesions and to negatives far from any annotation. At looser targets, what confidence can keep depends on the model; Gemma-4-12B keeps about a quarter of the slices at 10\%. Where we could measure it, confidence differs more between a model's two answers than between its right and wrong answers, so its apparent reliability depends on the population: the correlation reverses between glioma patients and healthy volunteers for all eight models. Mean confidence also changes little with lesion area. A model's statement about its own reliability should be validated like any other clinical signal: on the population where it will be used, before anyone acts on it.

\section{Acknowledgments}
Research was sponsored by the Army Research Laboratory and was accomplished under Cooperative Agreement Number W911NF-23-2-0224. The views and conclusions contained in this document are those of the authors and should not be interpreted as representing the official policies, either expressed or implied, of the Army Research Laboratory or the U.S. Government. The U.S. Government is authorized to reproduce and distribute reprints for Government purposes notwithstanding any copyright notation herein.
{
    \small
    \bibliographystyle{ieeenat_fullname}
    \bibliography{main}
}

\end{document}